\documentclass[10pt,twocolumn,letterpaper]{article}

\usepackage[pagenumbers]{wacv}
\usepackage{graphicx}
\usepackage{amsmath}
\usepackage{amssymb}
\usepackage{booktabs}
\usepackage{algorithm}
\usepackage{algpseudocode}
\usepackage[]{mdframed}
\usepackage{float}
\usepackage{ragged2e}

\def\eg{\emph{e.g}\onedot} 

\def\ie{\emph{i.e}\onedot}

\usepackage[pagebackref,breaklinks,colorlinks]{hyperref}
\usepackage[accsupp]{axessibility} %

\usepackage[capitalize]{cleveref}
\crefname{section}{Sec.}{Secs.}
\Crefname{section}{Section}{Sections}
\Crefname{table}{Table}{Tables}
\crefname{table}{Tab.}{Tabs.}

\def\confName{WACV}
\def\confYear{2025}

\begin{document}

\title{Enhancing Visual Classification using Comparative Descriptors}

\author{Hankyeol Lee$^{1}$, Gawon Seo$^{1}$, Wonseok Choi$^{1}$, Geunyoung Jung$^{1}$, Kyungwoo Song$^{2}$, Jiyoung Jung$^{1}$\\
$^{1}$Department of Artificial Intelligence, University of Seoul\\
$^{2}$Department of Applied Statistics, Yonsei University\\
{\tt\small \{leehk, gawon1224, wonseok.choi, gyjung975, jyjung\}@uos.ac.kr, kyungwoo.song@yonsei.ac.kr}
}

\maketitle

\begin{abstract}
The performance of vision-language models (VLMs), such as CLIP, in visual classification tasks, has been enhanced by leveraging semantic knowledge from large language models (LLMs), including GPT. Recent studies have shown that in zero-shot classification tasks, descriptors incorporating additional cues, high-level concepts, or even random characters often outperform those using only the category name. In many classification tasks, while the top-1 accuracy may be relatively low, the top-5 accuracy is often significantly higher. This gap implies that most misclassifications occur among a few similar classes, highlighting the model's difficulty in distinguishing between classes with subtle differences. To address this challenge, we introduce a novel concept of comparative descriptors. These descriptors emphasize the unique features of a target class against its most similar classes, enhancing differentiation. By generating and integrating these comparative descriptors into the classification framework, we refine the semantic focus and improve classification accuracy. An additional filtering process ensures that these descriptors are closer to the image embeddings in the CLIP space, further enhancing performance. Our approach demonstrates improved accuracy and robustness in visual classification tasks by addressing the specific challenge of subtle inter-class differences.
Code is available at \url{https://github.com/hk1ee/Comparative-CLIP}
\end{abstract}

\section{Introduction} %
\label{sec:intro}

The emergence of vision-language models (VLMs), such as CLIP~\cite{Radford-2021-CLIP}, has contributed significantly to the vast progress in the field of visual classification. The
standard zero-shot visual classification procedure – computing similarity between the query image and the embedded words for each category~\cite{Menon-2023-DCLIP}, then choosing the highest – has shown impressive performance on many popular benchmarks, including ImageNet~\cite{Deng-2009-ImageNet}. Comparing against the word that names a category was a reasonable place to start~\cite{Radford-2021-CLIP}.

While numerous studies have explored VLMs, recent research has shifted towards an innovative visual classification strategy~\cite{Menon-2023-DCLIP,Maniparambil-2023-EnhancingCW,Parashar-2023-Prompting,Han-2023-LLMsAV,Saha-2024-Improved} that enhances class representation by leveraging the semantic knowledge from large language models (LLMs) (\eg, GPT-3~\cite{Tom-2020-GPT3}).
Specifically, GPT-descriptor-extended CLIP (DCLIP)~\cite{Menon-2023-DCLIP} diverges from the existing method that relied solely on class names, \eg, ``A photo of \textbf{a Golden Retriever}.'' by incorporating additional information, \eg, ``A photo of a Golden Retriever, \textbf{which has golden fur}.'' termed a descriptor.
Employing such descriptors with additional cues in the class has led to improvements in classification performance as well as some explainability of the model. One can understand how the model classifies an image by checking the similarity of the image embedding with each descriptor. However, this approach has limitations mainly due to failure in descriptor creation. For example, GPT-3 occasionally produces descriptors about taste and smell in addition to vision, as well as descriptors with ambiguous words having multiple meanings, synthetic errors, or repetitive words~\cite{Menon-2023-DCLIP}. These descriptors diminish the image classification performance.

An interesting approach named WaffleCLIP~\cite{Roth-2023-Waffle} substitutes these LLM-generated descriptors with either random words \eg, ``foot loud'' or sequence of random characters \eg, ``jmhj, !J\#m'', which have absolutely no semantic relation to class names. Without any external models, classification accuracy remained comparable to that of semantically meaningful descriptors. Moreover, when high-level concepts are added, \eg, ``A photo of \textbf{an animal}: a Golden Retriever, which has jmhj, !J\#m.'', the performance substantially exceeds the previous descriptor-based approach (DCLIP)~\cite{Menon-2023-DCLIP}. This progression consequently triggered a reconsideration of the benefits of additional semantics introduced by LLM-generated descriptors~\cite{Roth-2023-Waffle}.
However, this approach has an inherent downside: the complete loss of explainability. While using LLM-generated descriptors improved classification accuracy and provided insights into the model's decision-making process~\cite{Menon-2023-DCLIP}, employing random descriptors to enhance performance significantly diminishes the interpretability of the model's decisions. %

\begin{figure*}[t]
  \centering
  \includegraphics[width=17cm]{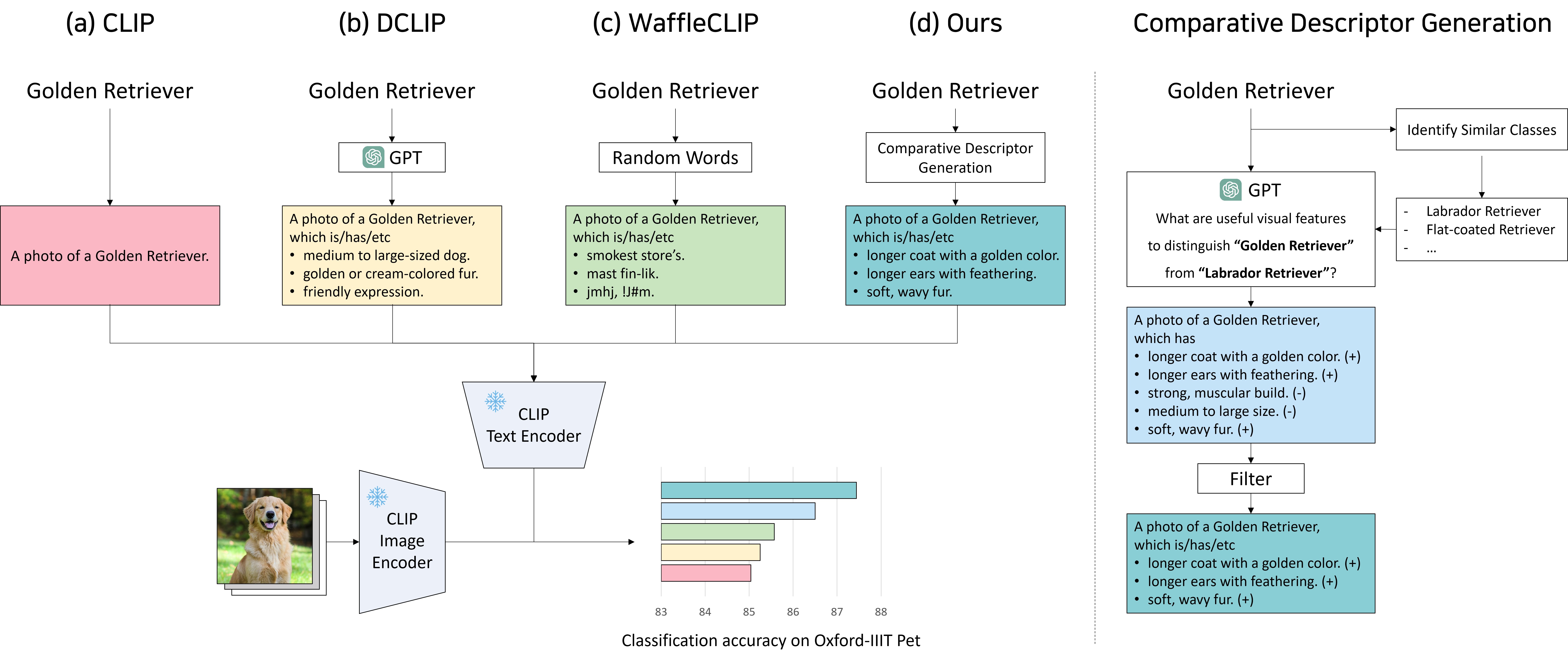}
  \caption{\textbf{Overview of our method and the baselines}(CLIP~\cite{Radford-2021-CLIP}, GPT-descriptor-extended CLIP (DCLIP)~\cite{Menon-2023-DCLIP}, and WaffleCLIP~\cite{Roth-2023-Waffle}). 
  (left) Comparison of our method with the baselines. Our proposed method outperforms the baselines on CLIP's image classification task.
  (right) A detailed overview of our method. Our method generates descriptors through comparison with semantically similar classes. Then the filtering process is applied to retain only descriptors that are useful for classification, significantly increasing the classification accuracy.}
  \label{fig:overview}
\end{figure*}

During our analysis of existing experiments across various visual classification tasks, we found an inspiring phenomenon: as shown in Table \hyperref[tab:motivation]{1}, while top-1 accuracy may be relatively low, top-5 accuracy is often significantly higher. For example, in zero-shot classification on the Caltech-UCSD Birds 200 dataset~\cite{Peter-2010-CUB200}, there's a notable difference: top-1 accuracy stands at 51.36\%, whereas top-5 accuracy is 83.48\%.
This gap implies that most misclassifications occur for a few similar classes out of the hundreds of classes in the dataset.
It also indicates that the model has difficulty distinguishing between classes with subtle differences, often confusing classes with common characteristics.
Based on this observation, we hypothesized that generating descriptors that emphasize the differences between the target class and its similar, frequently confused classes could enhance classification accuracy.

In this paper, we propose a novel approach to enhance the image classification performance of VLMs by generating \emph{comparative descriptors} using LLMs. Our approach consists of two steps.
Rather than directly prompting LLMs to generate descriptions for a specific class, we identify $n$ classes that are semantically similar in advance, by measuring the distances between the text features of each class name. We then request the LLMs to generate descriptions for the target class comparatively, highlighting its distinct features relative to these semantically similar classes.
For example, we query GPT, \texttt{`What are useful features for distinguishing a \{target class\} from a \{similar class\} in a photo?'} to generate comparative descriptors.

Next, we propose the following descriptor selection process, which we call the filtering process. This process is motivated by the observation that some descriptors generated by LLMs do not contribute to the classification and can even harm performance. Since VLMs and LLMs have different knowledge bases, some descriptors generated by LLMs may not be useful for classification in VLMs. Intuitively, removing these unhelpful descriptors would enhance classification accuracy. Specifically, we retain only the top-$k$ descriptors that exceed a certain amount of similarity to the mean image feature of each class. These remaining $k$ comparative descriptors effectively represent the target classes, distinguishing them well from their similar classes, thereby significantly enhancing the classification performance of VLMs.

In summary, our contributions are as follows:
1) We present a novel concept of comparative descriptors, which are visual descriptions emphasizing differences and unique features of the target class compared to semantically similar classes. 
This approach has reduced issues like ambiguity and the difficulty of understanding modalities, which are common problems associated with existing LLM-based descriptor generation methods.
2) We propose a simple yet effective filtering process that retains only class-specific descriptors contributing to image classification. By measuring the similarity of each descriptor to the mean image feature of its respective class, we ensure that only the most relevant and distinguishing descriptors are preserved. 
3) Our method significantly improves the image classification performance of VLMs (particularly CLIP)  across various datasets while still preserving the inherent interpretability of the model's decision.

Note that the classification leveraging comparative descriptors operates in a zero-shot setting, whereas the additional filtering process applies to a few-shot setting, as it involves calculating the mean image feature using a few examples for each class.
In the experiments section, we independently evaluate the impact of both the comparative descriptors and the filtering process on enhancing the classification performance of VLMs, demonstrating their respective contributions.

\begin{table*}[t]
\centering
\resizebox{\textwidth}{!}{%
\begin{tabular}{@{}c|ccccccccccccc|c@{}}
\toprule
\textbf{ViT-B/32} &
  \textbf{IN1k} &
  \textbf{IN1k-V2} &
  \textbf{Caltech} &
  \textbf{CIFAR} &
  \textbf{CUB} &
  \textbf{SAT} &
  \textbf{Places} &
  \textbf{Food} &
  \textbf{Pets} &
  \textbf{DTD} &
  \textbf{Flowers} &
  \textbf{Aircraft} &
  \textbf{Cars} &
  \textbf{Avg} \\ \midrule
Top-1 &
  62.04 &
  54.67 &
  78.39 &
  64.31 &
  51.36 &
  40.89 &
  39.11 &
  82.57 &
  85.04 &
  43.19 &
  62.97 &
  24.96 &
  58.62 &
  57.55 \\
Top-5 &
  87.67 &
  81.97 &
  91.31 &
  88.38 &
  83.48 &
  90.91 &
  69.71 &
  96.89 &
  97.36 &
  73.62 &
  84.42 &
  63.43 &
  89.84 &
  84.54 \\ \bottomrule
\end{tabular}}
\caption{\textbf{The top-1 and top-5 accuracy for image classification on various datasets using the CLIP ViT-B/32 model.} Even if the top-1 accuracy is relatively low, the top-5 accuracy is significantly high. This indicates that while the dataset contains many classes, the model can effectively capture coarse features but struggles to distinguish subtle differences between similar classes.}
\label{tab:motivation}
\end{table*}

\section{Related Work} %

\subsection{Vision-Language Models (VLMs)}
Vision-language models (VLMs)~\cite{Su-2020-VL-BERT,Radford-2021-CLIP,Ramesh-2021-DALL-E,Li-2022-BLIP,Li-2023-BLIP2,Liu-2023-LLaVA} are multi-modal architectures designed to process and integrate visual and linguistic information simultaneously. These models are capable of understanding both visual and textual data concurrently, learning the interactions between these two forms of information.
Motivated by the achievements of large language models (LLMs) like BERT~\cite{Devlin-2019-BERT} and GPT~\cite{Radford-2018-GPT1}, researchers have begun developing large-scale VLMs. These models, pre-trained on extensive datasets containing both images and text, show exceptional performance across various tasks.
Specifically, CLIP~\cite{Radford-2021-CLIP} sets a significant milestone in the advancement of VLMs. CLIP learns the general correlations between images and text, showcasing remarkable flexibility and achieving superior performance in a wide array of visual tasks. Notably, it has reached state-of-the-art results in zero-shot classification, demonstrating its ability to accurately recognize images from unseen categories. This highlights CLIP's proficiency in understanding and categorizing a broad spectrum of visual concepts, extending beyond predefined categories.

\subsection{Large Language Models (LLMs)}
Large language models (LLMs)~\cite{Radford-2018-GPT1,Radford-2019-GPT2,Tom-2020-GPT3,Devlin-2019-BERT,Raffel-2019-T5,Liu-2019-RoBERTa} have become foundational in the field of natural language processing. These powerful models excel at detecting patterns in vast amounts of text data, enabling them to tackle a diverse range of language-related tasks such as understanding language~\cite{Devlin-2019-BERT}, generating text~\cite{Radford-2019-GPT2}, translating languages~\cite{Vaswani-2017-Transformer}, and summarizing texts~\cite{Raffel-2019-T5}. The development of LLMs has significantly influenced VLM research. Leveraging the language comprehension capabilities and the knowledge learned from text data inherent in LLMs, VLMs have improved their ability to process and interpret the intricate relationships between images and text. This advancement has enhanced multi-modal learning~\cite{Lu-2019-ViLBERT}, facilitated the integration and expansion of knowledge across visual and textual domains~\cite{Tan-2019-LXMERT}, and led to the efficient integration and fine-tuning of VLMs using pre-trained LLMs~\cite{Zhong-2021-AdaptingLM}.

\subsection{Visual Classification of VLMs}
Zero-shot classification~\cite{Parades-2015-Embarrass,Ye-2017-ZeroShot} is a task that allows a model to classify new classes or objects that were not seen during the training process.
Recent studies leverage vision-language models (VLMs) for conducting zero-shot classification tasks.
Specifically, there is research focused on utilizing both class names and additional information in termed descriptors.
It is noteworthy that performance improves when using large language models (LLMs) to generate descriptors~\cite{Menon-2023-DCLIP,Maniparambil-2023-EnhancingCW,Han-2023-LLMsAV} or even employing random strings as descriptors~\cite{Roth-2023-Waffle}.
On the other hand, few-shot classification~\cite{Bateni-2020-ImrovedFS,Wang-2021-MTUNet,Zhang-2020-FewShotBinary,Vong-2022-FewShotNLR} is another branch that classifies new classes with minimal training data. This method is useful for real-world problems with limited data or labeling difficulties~\cite{Quan-2022-Label}.
Particularly, training-free few-shot classification~\cite{Zhang-2022-Tip-Adapter} is a type of few-shot learning technique that does not require training.
Although it requires no training, it shows significant performance improvements compared to zero-shot approaches by leveraging a few image-label pairs from the dataset.

\section{Method}
We first discuss image classification in CLIP and classification when using descriptors in Section~\hyperref[sec:Method-classification]{3.1}.
Next, We describe how to generate comparative descriptors in Section~\hyperref[sec:Method-step1]{3.2}.
Specifically, to proactively identify classes likely to be misclassified, we propose an efficient method to detect semantically similar classes. These similar classes can replace incorrect classes resulting from misclassification. We then generate a descriptor by querying the LLM with a comparison between the target class and similar classes.
Lastly, Section~\hyperref[sec:Method-step2]{3.3} covers the filtering process to keep only descriptors that contribute to classification.

\subsection{Image Classification with Descriptors}
\label{sec:Method-classification}
Given an image $x$ and target categories $C$, the visual classification procedure of CLIP~\cite{Radford-2021-CLIP} is defined as,
\begin{equation} 
\tilde{c} = \arg\max_{c \in C} s(\phi_{I}(x), \phi_{T}(f_c)) 
\end{equation} 
where $f_c$ denotes the prompt, \eg, ``\texttt{A photo of a Golden Retriever.}'', $\phi_{I}$ and $\phi_{T}$ represent the image and text encoders respectively, and $s(\cdot,\cdot)$ represents the similarity score or the distance between two feature vectors calculated by the dot product.

To improve classification performance, we add a class descriptor~\cite{Menon-2023-DCLIP} to the prompt $f_c$ as, %
``\texttt{A photo of a Golden Retriever, which has golden fur.}''
Given a set of descriptors \textit{$D_{c}$} for each class $c$, classification procedure is reformulated as,
\begin{equation}
\tilde{c} = \arg\max_{c \in C} \frac{1}{|D_{c}|} \sum_{d \in D_{c}} s(\phi_{I}(x), \phi_{T}(d))
\end{equation}

\subsection{Generating Comparative Descriptors} %
\label{sec:Method-step1}
\subsubsection{Identifying similar classes}
\label{sec:Method-step1-1}
To reduce misclassification, we identify classes similar to the target class by calculating the cosine similarity between the text features of their class names.
The reasoning behind this approach is that CLIP has knowledge of both modalities, vision, and language. Therefore, we assume that, when given only text, it can leverage its inherent visual knowledge to find similar classes.
The cosine similarity between the text features \( {\phi_{T}}(f_i)\) and \( {\phi_{T}}(f_j)\) of the classes \(f_i\) and \(f_j\) respectively is given by:

\begin{equation}
\cos(\phi_T(f_i), \phi_T(f_j)) = \frac{\phi_T(f_i) \cdot \phi_T(f_j)}{\|\phi_T(f_i)\| \|\phi_T(f_j)\|} \quad \forall i, j \in C
\end{equation}

where \(\|\cdot\|\) denotes the Euclidean norm of the vector.
We select $n$ similar classes having high cosine similarities with the target class.

\subsubsection{Querying LLMs}
\label{sec:Method-step1-2}
Next, our goal is to generate comparative descriptors which include distinguishing attributes between the target class and its semantically similar classes. The descriptors are generated using GPT-4o~\cite{Hurst-2024-GPT4o} with the prompts for the queries as follows:

\begin{mdframed}
{\small \textsf{Q: What are useful features for distinguishing a \{target class\} from a \{similar class\} in the photo?}}
\\
{\small \textsf{A: There are several useful visual features to tell the photo is a \{target class\}, not a \{similar class\}.}}
\end{mdframed}

The proposed approach addresses the limitations associated with LLM-generated descriptors, such as challenges in modality comprehension and word ambiguity, effectively reducing their impact.
This reduces the occurrence of misclassifications and thus improves the overall accuracy of classification tasks.

To effectively utilize in-context learning techniques with LLMs, we prepared ten sets of question-and-answer examples, following a consistent prompt template as shown above. For each instance of generating comparative descriptors, we randomly choose two sets from these ten. We arbitrarily select the target and similar classes from our datasets to create comparative descriptors using ChatGPT. This dynamic selection process, rather than using a fixed set of examples, introduces a broader range of contexts to the model.

\begin{algorithm}[b]
\caption{Filtering Process}
\label{algo1}
\begin{algorithmic}[1]
\For{$c \in \text{classes}$}
    \State $\text{\textit{Lower Bound}} \gets \text{min}(\text{cos}(\bar{\phi}_{Ic},\phi_{T}(f_c))$, 0.3)
    \For{$d \in D_{c}$}
        \State $\text{score} \gets \text{cos}(\bar{\phi}_{Ic},\phi_{T}(d))$
        \If{$\text{score} < \text{\textit{Lower Bound}}$}
            \State $\text{Discard}(d)$
        \EndIf
    \EndFor
    \State $D_{c} \gets \text{Select}(D_{c}, k)$
    
    \If{$D_{c}\text{ is empty}$}
        \State $D_{c} \gets f_c$
    \EndIf
\EndFor
\end{algorithmic}
\end{algorithm}

\begin{table*}[t]
\resizebox{\textwidth}{!}{%
\begin{tabular}{@{}l|ccccccccccccc|c@{}}
\toprule
\multicolumn{1}{c|}{\textbf{ViT-B/32}} &
  \textbf{IN1k} &
  \textbf{IN1k-V2} &
  \textbf{Caltech} &
  \textbf{CIFAR} &
  \textbf{CUB} &
  \textbf{SAT} &
  \textbf{Places} &
  \textbf{Food} &
  \textbf{Pets} &
  \textbf{DTD} &
  \textbf{Flowers} &
  \textbf{Aircraft} &
  \textbf{Cars} &
  \textbf{Avg} \\ \midrule
CLIP             & 62.04 & 54.67 & 78.39  & 64.31  & 51.36 & 40.89 & 39.11 & 82.57 & 85.04 & 43.19 & 62.97 & 24.96 & 58.62 & 57.55 \\
DCLIP            & 63.66 & 56.28 & 81.23  & 64.94  & 53.61 & 41.37 & 41.64 & 83.06 & 85.25 & 44.26 & 66.63 & 26.67 & 59.08 & 59.05 \\
WaffleCLIP       & 63.32 & 55.97 & 81.06  & 65.44  & 52.47 & 43.66 & 40.67 & 82.84 & 85.48 & 42.93 & 66.32 & 25.70 & 58.82 & 58.82 \\
Ours             & 64.02 & 56.71 & 82.24  & 65.69  & 54.09 & 42.92 & 42.22 & 83.81 & 87.41 & 46.54 & 67.07 & 27.57 & 59.30 & 59.97 \\
Ours + Filtering & 65.66 & 57.56 & 83.55 & 64.81 & 56.41 & 51.08 & 44.18 & 84.62 & 87.04 & 53.78 & 73.43 & 28.61 & 60.15 & 62.37 \\ \bottomrule
\end{tabular}
}
\caption{\textbf{Results of the image classification.} Comparison of classification results using our method with CLIP \cite{Radford-2021-CLIP}, DCLIP \cite{Menon-2023-DCLIP}, and WaffleCLIP \cite{Roth-2023-Waffle}. Our method outperforms the baseline performances across most evaluated datasets. Results for ViT-L/14 and ResNet50 are available in the supplementary material.}
\label{tab:image classification result}
\end{table*}

\begin{table*}[t]
\centering
\resizebox{\textwidth}{!}{%
\begin{tabular}{@{}c|ccccccccccccc|c@{}}
\toprule
\textbf{ViT-B/32} &
  \textbf{IN1k} &
  \textbf{IN1k-V2} &
  \textbf{Caltech} &
  \textbf{CIFAR} &
  \textbf{CUB} &
  \textbf{SAT} &
  \textbf{Places} &
  \textbf{Food} &
  \textbf{Pets} &
  \textbf{DTD} &
  \textbf{Flowers} &
  \textbf{Aircraft} &
  \textbf{Cars} &
  \textbf{Avg} \\ \midrule
DCLIP &
  22.53 &
  19.69 &
  48.92 &
  35.04 &
  4.45 &
  38.43 &
  26.22 &
  37.03 &
  17.06 &
  25.48 &
  12.67 &
  7.98 &
  13.17 &
  23.74 \\
Ours &
  31.57 &
  27.54 &
  60.60 &
  41.07 &
  18.62 &
  46.17 &
  33.37 &
  51.54 &
  35.76 &
  35.69 &
  25.09 &
  14.97 &
  28.73 &
  34.67 \\ \bottomrule
\end{tabular}%
}
\caption{\textbf{Image classification results without class labels}. To evaluate the quality of the descriptor itself, we perform image classification after removing class labels and prefixes, using only the descriptor as text input. In this experiment, our method outperforms the existing methods. This means that our descriptors contribute significantly to the improvement of classification performance.}
\label{tab:without classname}
\end{table*}

\subsection{Filtering}
\label{sec:Method-step2}

In the final stage, we filter the generated descriptors to include only those that contribute to classification.
Initially, we compute the mean image feature for each class by selecting a few images of the class from the dataset, making our method applicable even in scenarios with limited image availability.
Note that the mean image feature is utilized solely during this filtering phase.
We calculate the cosine similarity between the mean image feature of each class and every descriptor, deriving similarity scores. From these, we retain only the top-$k$ descriptors based on their scores, discarding the rest.

We employ a lower bound for similarity scores in this process, set as the cosine similarity between the mean image feature and a CLIP-style text prompt (\eg, ``A photo of a \{class\}.''). Descriptors falling below this threshold are excluded. If no descriptors for a class exceed this lower bound, classification proceeds using only the CLIP-style text prompt, without additional descriptors. For detailed steps of this filtering process, refer to the pseudo-code presented in Algorithm \hyperref[algo1]{1}.
\section{Experiments}
First, we provide experimental details in Section \hyperref[sec:Experimental-details]{4.1}.
We then assess the effectiveness of our method through image classification in Section \hyperref[sec:Experiment-classification]{4.2}, demonstrating its superior performance compared to existing methods.
Next, in Section \hyperref[sec:Experiment-only-descriptor]{4.3}, we perform image classification using only descriptors as text input and show that our method generates high-quality descriptors.
Additionally, we show that our method is not solely dependent on the number of descriptors by comparing it to the baseline after equalizing the number of descriptors in Section \hyperref[sec:Experiment-equal-number]{4.4}.
We propose and evaluate a few-shot filtering process in Section \hyperref[sec:Experiment-few-shot]{4.5}, highlighting its practical utility even with limited data.
Also, we apply our filtering process to the existing method to show the effectiveness of this process.
We show the explainability, which is an advantage of using the LLM-generated descriptor in Section \hyperref[sec:Experiment-explainability]{4.6}.
Finally, we discuss some limitations of our approach in Section \hyperref[sec:Experiment-limitations]{4.7}, providing a foundation for future research and improvements.

\subsection{Experimental Details}
\label{sec:Experimental-details}
In all experiments, we utilize CLIP~\cite{Radford-2021-CLIP} as the underlying VLM. Furthermore, since our method is training-free, all experiments were conducted using a single NVIDIA 3090 GPU.

\subsubsection{Hyperparameters}
In our experiment, we use two hyperparameters: the number of similar classes per each class, denoted as \emph{n}, and the maximum number of descriptors to retain after filtering, denoted as \emph{k}. We empirically choose one of the values between 5 and 20 for both \emph{n} and \emph{k}. The hyperparameters used for each dataset are shown in supplementary material.

\subsubsection{GPT}
In the baselines~\cite{Menon-2023-DCLIP,Roth-2023-Waffle}, GPT-3~\cite{Tom-2020-GPT3} text-davinci-003 was employed as the LLM to generate descriptors, but this version has been deprecated.
Therefore, to ensure a fair comparison, we re-conducted all experiments for both the baselines and our method using GPT-4o~\cite{Hurst-2024-GPT4o}.
Note that we did not change any of the baselines' settings, except for the GPT version. To demonstrate the fairness of our experiments, we disclose all responses.

\subsubsection{Datasets}
We conduct experiments on 13 datasets, including the 11 datasets tested in the baselines\cite{Menon-2023-DCLIP, Roth-2023-Waffle}: ImageNet (IN1k)~\cite{Deng-2009-ImageNet}, ImageNetV2 (IN1k-V2)~\cite{Kornblith-2018-ImageNetv2}, CUB200-2011 (CUB)~\cite{Peter-2010-CUB200}, EuroSAT (SAT)~\cite{Helber-2019-EuroSAT}, Places365 (Places)~\cite{Zhou-2018-Places365}, Food101 (Food)~\cite{Bossard-2014-Food101}, Oxford-IIIT Pet (Pets)~\cite{Parkhi-2012-Pets}, Describable Textures (DTD)~\cite{Cimpoi-2014-DTD}, Flowers102 (Flowers)~\cite{Nilsback-2008-Flowers102}, FGVCAircraft (Aircraft)~\cite{Maji-2013-FGVCAircraft}, and Stanford Cars (Cars)~\cite{Krause-2013-StanfordCars}, along with 2 additional datasets: Caltech256 (Caltech) \cite{Griffin-2007-Caltech256} and CIFAR100 (CIFAR) \cite{Krizhevsky-2009-LearningML}.
Note that ImageNetV2 is an expansion of the validation set designed to evaluate the generalization ability of ImageNet, sharing the same set of classes and not including a new training set. Therefore, we use the mean image feature of ImageNet during the filtering process for ImageNetV2.

\subsection{Image Classification}
\label{sec:Experiment-classification}
We assess the effectiveness of our method by performing image classification on various datasets, each of which contains a wide range of domains.
We then compare these results to those obtained by existing methods, as shown in Table \hyperref[tab:image classification result]{2}.

\vspace{3mm}

Since our method generates descriptors by comparing similar classes, it is straightforward to perform well on domain-specific datasets such as Flowers ~\cite{Nilsback-2008-Flowers102} and Cars ~\cite{Krause-2013-StanfordCars} which contain only images of flowers and cars, respectively. However, its performance on multi-domain datasets might be questionable. To address this concern, we present additional experimental results on two multi-domain datasets: Caltech \cite{Griffin-2007-Caltech256} and CIFAR \cite{Krizhevsky-2009-LearningML}, which were not covered in the baselines \cite{Menon-2023-DCLIP, Roth-2023-Waffle}.
Our method outperforms the baselines across most evaluated datasets, regardless of the domain. This superior performance demonstrates the validity and robustness of our approach.

\begin{figure}[t]
  \centering
  \includegraphics[width=8cm]{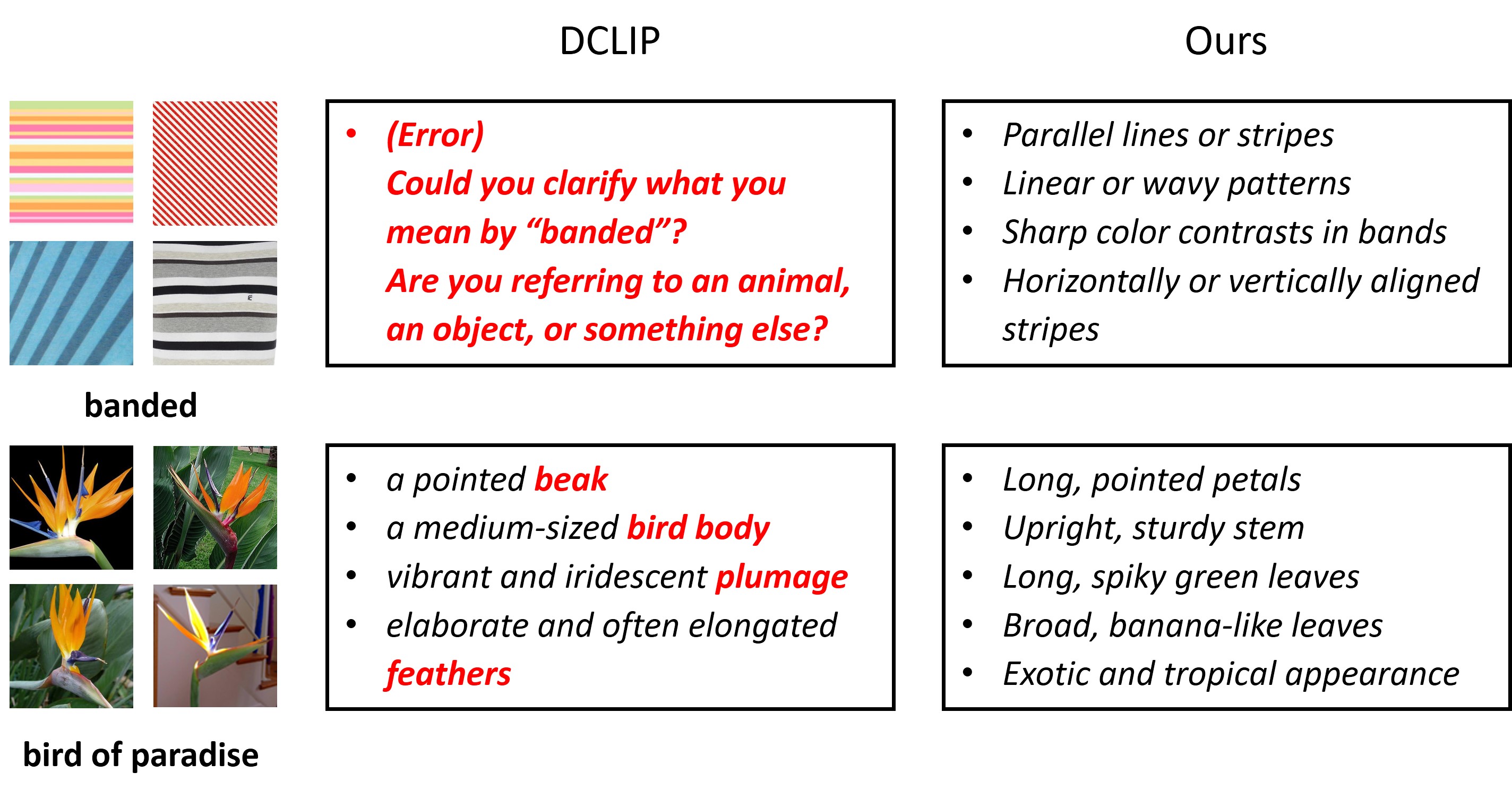}
  \caption{\textbf{Addressing ambiguity in descriptor generation.} We compare the descriptors generated by our method and the DCLIP \cite{Menon-2023-DCLIP} method on the (top) Describable Texture Dataset (DTD)~\cite{Cimpoi-2014-DTD} and (bottom) Flowers102 (Flowers)~\cite{Nilsback-2008-Flowers102} dataset.
  DCLIP method failed to generate descriptors due to ambiguity (\eg the word \textit{banded} refers to both texture and a species of snake), and in some cases generated descriptors that were unrelated to the class.
  On the other hand, our method not only avoided failures but also enriched in context. This difference leads to a significant improvement in classification accuracy.
  }
  \label{fig:dtd}
\end{figure}

In particular, the classification accuracy on the Describable Texture Dataset (DTD)~\cite{Cimpoi-2014-DTD} was significantly improved compared to existing methods.
When we generate descriptors using the existing method~\cite{Menon-2023-DCLIP}, we frequently encounter failures due to the ambiguity of words. This leads to the production of descriptors irrelevant to the intended class and lacking sufficient detail.
Specifically, with GPT-4o~\cite{Hurst-2024-GPT4o}, descriptors are sometimes not created at all if the class is ambiguous.
In contrast, we can make precise and relevant queries by generating descriptors using our comparison-based method.
This approach reduces the occurrence of errors and produces context-rich descriptors, ultimately increasing classification accuracy.
To offer further insight into these results, we compare the descriptors generated by DCLIP and our approach. See Figure \hyperref[fig:dtd]{2} for examples.

\subsection{Image Classification with Descriptors Only} %
\label{sec:Experiment-only-descriptor}
To evaluate the quality of the descriptor, we modify the form of the input text. Instead of using the format ``\texttt{A photo of a \{class\}, which is/has/etc \{descriptor\}.}" in Section \hyperref[sec:Experiment-classification]{4.2}, we remove the class name and prefix, using only the descriptor as the text input, \ie ``\texttt{\{descriptor\}}", for classification.
The results of this experiment are shown in Table \hyperref[tab:without classname]{3}. Our method achieves a much higher accuracy for all evaluated datasets than the existing method.

\begin{figure}[t]
  \centering
  \includegraphics[width=8cm]{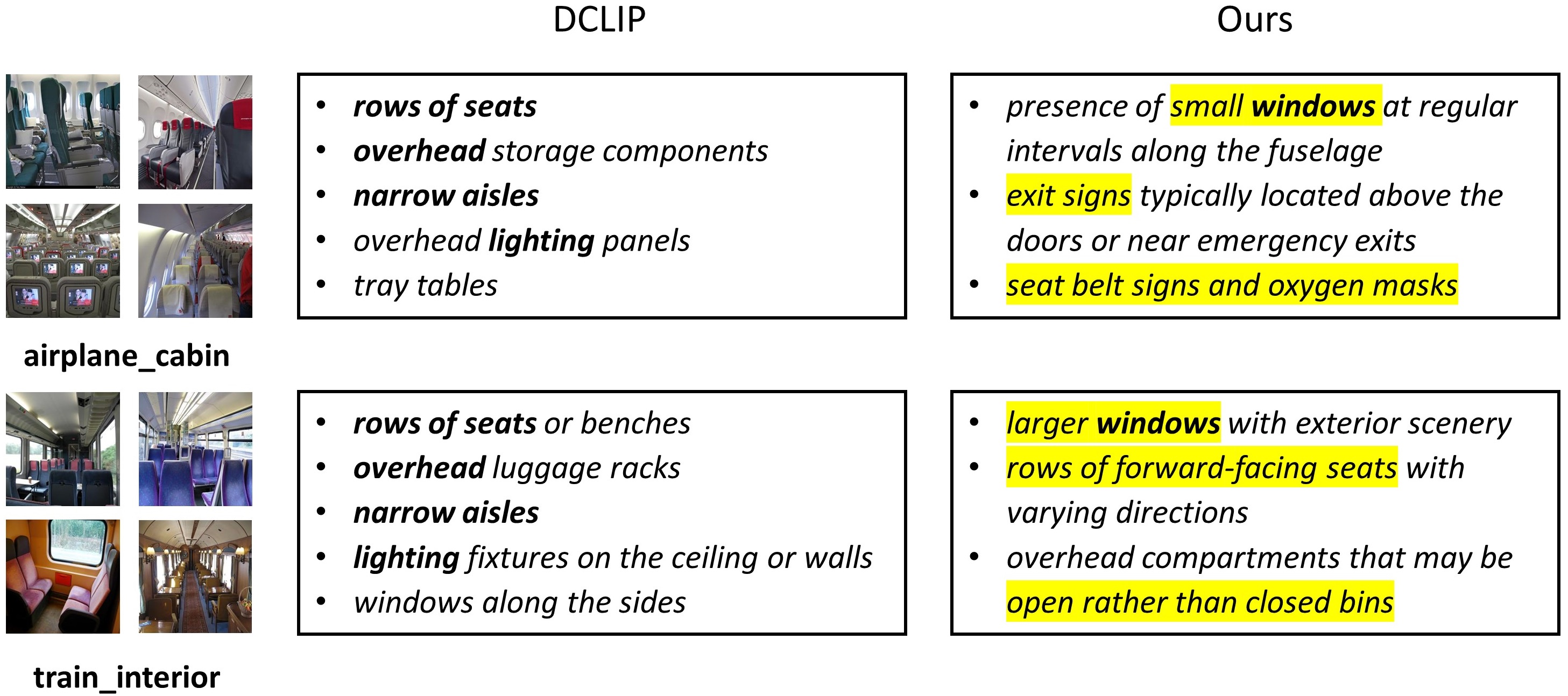}
  \caption{\textbf{Comparison of generated descriptors between similar classes}. %
  Descriptors were generated for similar classes in the Places dataset.
  Bold text indicates an attribute that appears in both classes and highlighted text indicates a distinct attribute.
  The descriptors generated by DCLIP have semantically equivalent descriptors between the target class and its similar class. In contrast, our method minimizes semantically analogous descriptors, adds distinct features, and provides more detailed explanations. As a result, this makes it easier to distinguish between similar classes.
  }
  \label{fig:similar}
\end{figure}

One of the reasons for these results can be found in the reduction of common attributes in the descriptors.
The descriptors generated by the existing method are likely to contain common attributes. In contrast, our method generates descriptors by comparison, minimizing the generation of common attributes and allowing us to obtain distinct attributes.
To be more specific, no matter how well a descriptor describes a class, if a semantically analogous descriptor exists for another class, this descriptor will not help distinguish between them.
For better understanding, we present the examples in Figure \hyperref[fig:similar]{3}.
Consequently, this result demonstrates that the descriptors generated by our method have a much higher contribution to the classification.

\begin{table}[t]
\centering
\small
\begin{tabular}{@{}c|cc@{}}
\toprule
\textbf{ViT-B/32} & \textbf{Random (k=5)} & \textbf{Filtering (k=5)} \\ \midrule
DCLIP             & 58.31                 & 59.68                    \\
Ours              & 58.66                 & 61.97                    \\ \bottomrule
\end{tabular}%
\caption{\textbf{Image classification results with an equal number of descriptors.} We reduce the number of descriptors in two ways to make them equal, and both cases show better performance when using our descriptors. This demonstrates that our method does not depend solely on the number of descriptors.}
\label{tab:equal}
\end{table}

\begin{table*}[t]
\centering
\resizebox{\textwidth}{!}{%
\begin{tabular}{@{}c|ccccccccccccc|c@{}}
\toprule
\textbf{Shot} &
  \textbf{IN1k} &
  \textbf{IN1k-V2} &
  \textbf{Caltech} &
  \textbf{CIFAR} &
  \textbf{CUB} &
  \textbf{SAT} &
  \textbf{Places} &
  \textbf{Food} &
  \textbf{Pets} &
  \textbf{DTD} &
  \textbf{Flowers} &
  \textbf{Aircraft} &
  \textbf{Cars} &
  \textbf{Avg} \\ \midrule
1 &
  63.97 &
  56.23 &
  82.67 &
  64.38 &
  54.59 &
  48.88 &
  41.91 &
  84.03 &
  86.56 &
  50.94 &
  72.34 &
  28.03 &
  56.92 &
  60.88 \\
2 &
  64.54 &
  56.59 &
  83.16 &
  64.45 &
  55.41 &
  50.78 &
  42.56 &
  84.25 &
  86.77 &
  52.09 &
  72.80 &
  28.03 &
  58.53 &
  61.54 \\
4 &
  65.10 &
  57.08 &
  83.26 &
  64.57 &
  56.11 &
  50.49 &
  43.29 &
  84.34 &
  86.93 &
  52.88 &
  72.97 &
  28.34 &
  59.29 &
  61.90 \\
8 &
  65.36 &
  57.33 &
  83.49 &
  64.78 &
  56.36 &
  50.39 &
  43.79 &
  84.44 &
  86.82 &
  53.33 &
  \textbf{73.43} &
  \textbf{28.61} &
  59.85 &
  62.15 \\
16 &
  65.52 &
  57.44 &
  83.43 &
  64.76 &
  \textbf{56.41} &
  51.05 &
  44.09 &
  84.51 &
  \textbf{87.04} &
  53.70 &
  - &
  28.59 &
  \textbf{60.15} &
  - \\
32 &
  65.65 &
  \textbf{57.56} &
  \textbf{83.55} &
  \textbf{64.81} &
  - &
  \textbf{51.08} &
  44.16 &
  84.52 &
  86.92 &
  \textbf{53.78} &
  - &
  28.48 &
  - &
  - \\
64 &
  \textbf{65.66} &
  57.55 &
  - &
  64.71 &
  - &
  50.93 &
  \textbf{44.18} &
  \textbf{84.62} &
  86.88 &
  - &
  - &
  - &
  - &
  - \\ \bottomrule
\end{tabular}%
}
\caption{\textbf{Image classification results with few-shot filtering applied.} Our filtering method shows superior results even when using a few images. This indicates that our method can be useful even in situations with limited data.}
\label{tab:few-shot}
\end{table*}

\begin{table*}[t]
\centering
\resizebox{\textwidth}{!}{%
\begin{tabular}{@{}l|ccccccccccccc|c@{}}
\toprule
\multicolumn{1}{c|}{\textbf{ViT-B/32}} &
  \textbf{IN1k} &
  \textbf{IN1k-V2} &
  \textbf{Caltech} &
  \textbf{CIFAR} &
  \textbf{CUB} &
  \textbf{SAT} &
  \textbf{Places} &
  \textbf{Food} &
  \textbf{Pets} &
  \textbf{DTD} &
  \textbf{Flowers} &
  \textbf{Aircraft} &
  \textbf{Cars} &
  \textbf{Avg} \\ \midrule
DCLIP &
  63.66 &
  56.28 &
  81.23 &
  64.94 &
  53.61 &
  41.37 &
  41.64 &
  83.06 &
  85.25 &
  44.26 &
  66.63 &
  26.67 &
  59.08 &
  59.05 \\
+ Filtering &
  64.29 &
  56.42 &
  81.22 &
  63.83 &
  54.24 &
  42.85 &
  42.57 &
  83.61 &
  85.03 &
  46.06 &
  67.87 &
  28.20 &
  59.59 &
  59.68 \\ \bottomrule
\end{tabular}%
}
\caption{\textbf{Image classification results when filtering is applied to the existing method \cite{Menon-2023-DCLIP}}. Performance improved in most datasets. This demonstrates that our proposed filtering method is effective not only for descriptors generated by our method but also for those generated by various other methods.}
\vspace{-3mm}
\label{tab:filtered DCLIP}
\end{table*}

\subsection{Evaluation with Equal Number of Descriptors}
\label{sec:Experiment-equal-number}
WaffleCLIP ~\cite{Roth-2023-Waffle} pointed out that the number of descriptors can affect image classification accuracy.
Our method identifies \emph{n} similar classes for one target class and appends all the descriptors generated from \emph{n} comparisons, it inevitably uses more descriptors than the baselines.
Since it may be questionable that the performance improvement with our method is due to the increased number of descriptors, we perform an additional experiment with the same number of descriptors in the baseline and our approach.

We try both random selection and filtering to equalize the number of descriptors. The number of descriptors, \emph{k}, is set to 5 for both approaches. We run both five times with different seeds or different images, and the average accuracy is shown in Table \hyperref[tab:equal]{4}.
Our method performs the baseline on average for both approaches. This demonstrates that the performance of our method is not solely dependent on the number of descriptors, but also indicates that it possesses better quality on its own.

\subsection{Few-Shot Filtering}
\label{sec:Experiment-few-shot}
We propose a simple, training-free filtering process to retain only descriptors close to the given class and contribute to the classification.
We present the classification accuracy obtained through this filtering process in the last row of Table \hyperref[tab:image classification result]{2}, and show the variation in performance depending on the number of images used in Table \hyperref[tab:few-shot]{5}.
We repeat the experiment five times using different images to get reliable results and present the average accuracy.
For most of the evaluated datasets, filtering using only one or two images also results in significant performance improvements. This demonstrates that our filtering can be used practically even in limited data situations.

Additionally, to demonstrate the validity of our proposed filtering process, we apply it to descriptors generated with DCLIP~\cite{Menon-2023-DCLIP} and compare the results before and after filtering.
As shown in Table \hyperref[tab:filtered DCLIP]{6}, the use of filtering generally leads to an increase in accuracy.

However, the performance improvement is relatively low compared to applying filtering to the descriptors produced by our method, and for some datasets, we even observe a slight decrease in accuracy.
This is because there are fewer descriptors available for filtering. Our filtering method aims to identify descriptors that are closely related to the class and useful for classification from a large pool of descriptors. If the number of these descriptors is limited, the filtering may be less effective.

\subsection{Explainability}
\label{sec:Experiment-explainability}
Our approach leverages descriptors generated by LLMs for image classification. Similar to DCLIP \cite{Menon-2023-DCLIP}, our method not only allows the model to make precise decisions but also provides an explanation of those decisions and the reasons behind them.
This capability significantly enhances the explainability of the model and provides deeper insights into its operational process.

\begin{figure*}[t]
  \centering
  \includegraphics[width=17cm]{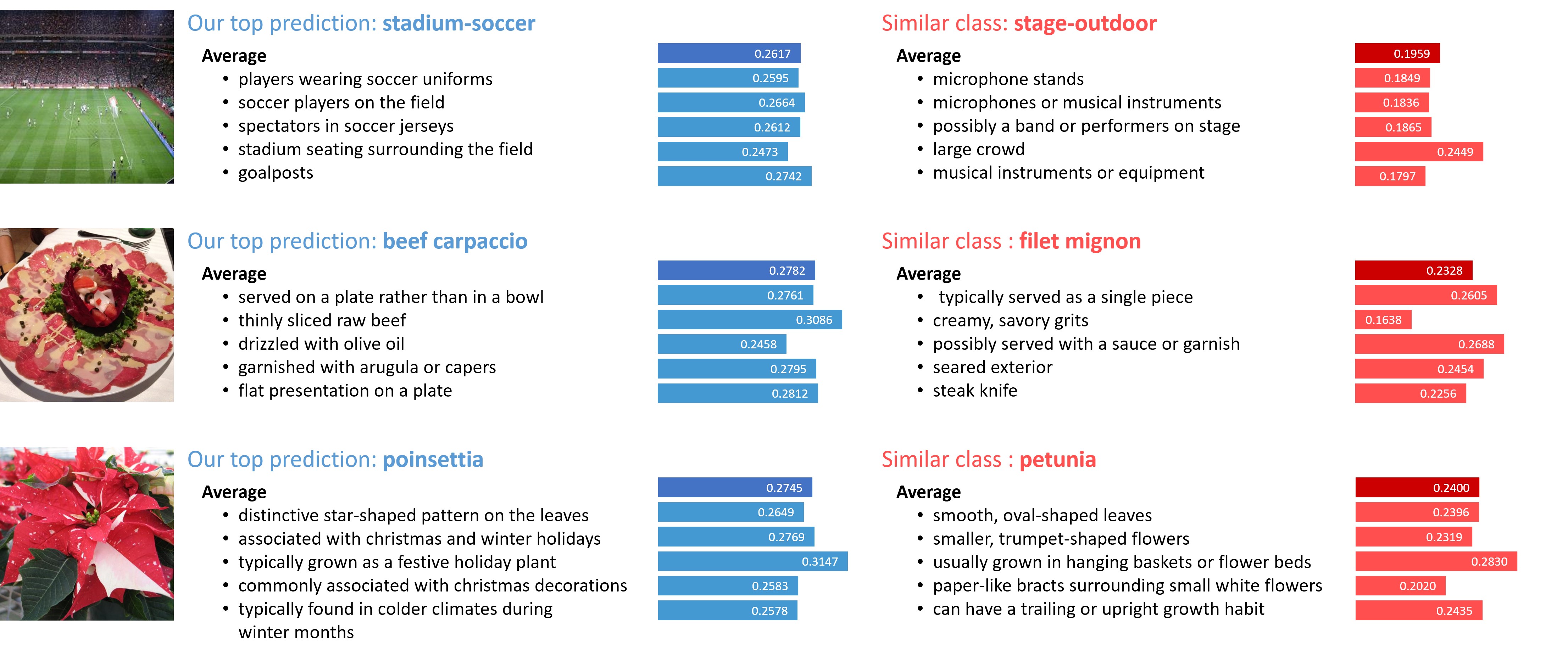}
  \caption{\textbf{Examples of explainability.} (left) We show examples of decisions and justifications through our model. (right) We present the descriptors of each target class and the similar class within the datasets (from the top, Places365~\cite{Zhou-2018-Places365}, Food101~\cite{Bossard-2014-Food101}, Flowers102~\cite{Nilsback-2008-Flowers102}). The chart shows the similarity score between the descriptor and the image in the CLIP latent space.
  The higher the similarity score, the greater the influence of the descriptor on the decision-making process. The descriptor with the highest average similarity becomes the model's classification prediction.
  }
  \label{fig:explainability}
\end{figure*}

As shown in Figure \hyperref[fig:explainability]{4}, we analyze by selecting a class and its similar counterpart from the dataset.
Despite their resemblance, we find that the similarity scores of the target and similar classes show a huge gap.
This gap enables the precise classification between the images of similar classes.
Moreover, by comparing similarity scores, we can identify which features in the image contribute more to classification. This provides a clearer understanding of the reasoning process and enhances the explainability of our model.

\subsection{Limitations}
\label{sec:Experiment-limitations}
In this section, we provide a detailed analysis of the limitations involved in our approach. This analysis will be useful in guiding future improvements and research directions.

We encounter an interesting problem while generating descriptors with LLMs for certain datasets.
FGVCAircraft (Aircraft)~\cite{Maji-2013-FGVCAircraft} and Stanford Cars (Cars)~\cite{Krause-2013-StanfordCars} are datasets that contain different types of aircraft and cars, respectively.
When querying on these datasets, we find some attributes that contain visual information but are invisible.
Think of the seating arrangement in aircraft or the engine in cars: these are visual components, but they are hard to observe due to the limited availability of images of aircraft and car interiors in the datasets.
This resulted in less improvement in classification accuracy on these datasets compared to other datasets.

During our analysis of the experimental results, we discovered that the use of filtering can lead to a lack of diversity in the descriptors.
More specifically, this phenomenon means that some filtered descriptors are semantically almost identical, but only slightly different in form (\eg, ``short, rounded ear'' and ``small, rounded ears'').
We assume that this phenomenon occurs because our filtering process only considers similarity scores. If an attribute has the highest similarity to the mean image feature of the class, it is reasonable to expect that a descriptor with a similar meaning and almost identical form would also have a high similarity.
We will address this issue in future work by considering more factors in the filtering process.

\section{Conclusion}
In this work, we presented an innovative strategy for enhancing the image classification capabilities of vision-language models (VLMs) by generating comparative descriptors. By comparing the target class with semantically similar classes, our method effectively minimizes ambiguity and enhances the model's ability to distinguish closely related classes.
Coupled with a filtering process, our approach substantially boosts classification accuracy across various datasets, demonstrating notable improvements over existing baselines, even with limited image-label pairs. Integrating comparative descriptors with our filtering mechanism significantly advances image classification performance.
This novel method not only improves the performance metrics of VLMs, particularly in models like CLIP ~\cite{Radford-2021-CLIP} but also enriches the interpretability of the results, offering a more intuitive understanding of the model's decision-making process.

\clearpage %

{\small
\bibliographystyle{ieee_fullname}
\bibliography{egbib}
}

\end{document}